\documentclass[runningheads]{llncs}
\pdfoutput=1 
\usepackage{graphicx}
\usepackage{amsmath}
\usepackage{amssymb}

\usepackage[pagebackref=false,breaklinks=true,letterpaper=true,colorlinks,bookmarks=false]{hyperref}

\mathchardef\mhyphen="2D
\def\etal{\emph{et al}.}

\begin{document}

\title{Multi-Person Absolute 3D Human Pose Estimation with Weak Depth Supervision}

\author{M\'{a}rton V\'{e}ges \and
Andr\'{a}s L\H{o}rincz}

\authorrunning{M. V\'{e}ges et al.}
\institute{Eotvos Lorand University, Budapest, Hungary \\
\email{\{vegesm,lorincz\}@inf.elte.hu}}

\maketitle

\begin{abstract}
In 3D human pose estimation one of the biggest problems is the lack of large, diverse datasets. This is especially true for multi-person 3D pose estimation, where, to our knowledge, there are only machine generated annotations available for training. To mitigate this issue, we introduce a network that can be trained with additional RGB\nobreakdash-D images in a weakly supervised fashion. Due to the existence of cheap sensors, videos with depth maps are widely available, and our method can exploit a large, unannotated dataset. Our algorithm is a monocular, multi-person, absolute pose estimator. We evaluate the algorithm on several benchmarks, showing a consistent improvement in error rates. Also, our model achieves state-of-the-art results on the MuPoTS-3D dataset by a considerable margin. Our code will be publicly available\footnote{\url{https://github.com/vegesm/wdspose}}.
\end{abstract}

\section{Introduction}\label{sec:intro}

While the initial focus in 3D pose estimation was on single pose estimators \cite{zhou2017,fang2018posegrammar,helge_geometry-aware,Hossain2017temporal,3dbaseline,mehta}, recently multi-pose methods also started to appear \cite{veges2019depthpose,moon2019camdistance,zanfir2018multipose,dabral2019multiperson}. Usually, in single-pose detection the problem is simplified to relative pose prediction, that is, joint coordinates are estimated relative to a root joint. However, in a multi-pose setting this could be insufficient for downstream tasks. For instance, to detect proximity, locations are needed as well. The combined prediction of the relative pose and the location of the person is called absolute pose estimation \cite{veges2019depthpose}. Here, the coordinates of the joints are predicted in a coordinate system relative to the camera, instead of the hip.

The largest obstacle in 3D pose estimation is the lack of large, diverse datasets. Annotations require special equipment, thus large 3D pose datasets are restricted to a studio setting \cite{humaneva,h36m}. This is especially true for multi-pose estimation, where, to our knowledge, the only non-synthetic multi-pose 3D database for training is Panoptic \cite{panoptic}, and it only has approximate 3D joint coordinates.

To overcome this issue, various methods using different kind of auxiliary datasets were proposed: 2D pose datasets \cite{zhou2017}, synthetic poses \cite{llcs} or 3D poses generated from multi-view cameras  \cite{kocabas2019epipolar}.
In our work, we focus on RGB\nobreakdash-D videos. Videos with depth maps have advantages. First, it is not required to manually annotate additional data, unlike 2D pose datasets. Second, RGB-D videos can be recorded with a single camera, while multi-camera setups require special hardware with synchronization. The accessibility of the Kinect resulted in a multitude of RGB-D datasets taken in diverse environments \cite{firman2016rgbddatasets}. Specifically, we use Panoptic \cite{panoptic} as a weak supervisory signal.

However, depth maps do not provide an accurate representation of the 3D body joint location. Keypoints are easily occluded by other objects, or by the person him/herself. To solve the problem of occlusion we propose a pose decoder network that reconstructs the depth map at the joint locations.
Then, the output of the network can be compared with the ground truth depth and the errors backpropagated.

To summarize, we propose a method that exploits RGB\nobreakdash-D datasets to improve absolute 3D pose estimation. The RGB\nobreakdash-D datasets do not require further annotations, only camera calibration matrices are needed. We tested our method on the Panoptic and MuPo-TS datasets, improving previous results and achieving state-of-the-art results in absolute pose estimation.

\section{Related work}

\textbf{3D Pose Estimation} Although various approaches were used for 3D pose estimation, such as dictionary based methods \cite{ramakrishna2012dictionary,zhou2015dict} or conditional random fields \cite{belagiannis2014pictorialpose}, recently state-of-the-art results are dominated by deep learning based algorithms  \cite{lichan2015humanpose,3dbaseline,gorog,rogez2017lcrnet,llcs,integralPose}. Methods include regressing 3D joints directly \cite{lichan2015humanpose}, using volumetric heatmaps  \cite{gorog} or including soft-argmax layers \cite{integralPose}.

A common approach is to split the 3D estimation into two steps: first estimating 2D joint coordinates with an off-the-shelf pose detector, then lifting the 2D coordinates into 3D \cite{fang2018posegrammar,li2019mdn,3dbaseline,veges2018siamese}. In \cite{veges2018siamese}, an equivariant network is built to decrease the overfitting to cameras in the training set. Li and Lee \cite{li2019mdn} predict a distribution for each joint instead of a point estimation to produce uncertainty estimations.

\textbf{Weak- and semi-supervised approaches} The lack of diverse datasets for 3D pose estimation led to interest in weak- and self-supervised approaches. For instance,  depthwise ordinal ranking of joints helps training \cite{pavlakos2018ordinal}. Datasets with 2D annotations can be used as well, via a reprojection loss \cite{zhou2017}.
Another approach is to add adversarial losses. Drover \etal ~\cite{drover_2dto3d} backproject rotated poses into 2D and regularize them with a discriminator trained on real 2D poses.

Additionally, multi-camera setup can effectively decrease the required amount of training data \cite{kocabas2019epipolar,helge_geometry-aware}. Kocabas \etal \cite{kocabas2019epipolar} uses epipolar geometry to generate 3D ground truth data from unannotated multi-view images.

\textbf{Absolute pose estimation} Unlike the above approaches, absolute pose estimation predicts not only the root-relative pose, but the location of the person as well. One approach is to find an optimal translation that minimizes the reprojection error \cite{mehta,zanfir2018smpl3dpose} to 2D. However, this needs the 2D pose to be estimated along the 3D pose. Additionally, the prediction is done in a separate stage preventing end-to-end training. Moon \etal \cite{moon2019camdistance} predicts the location of the root joint directly, using a separate network. In contrast, in the method of \cite{veges2019depthpose}, the location prediction and relative pose estimation share the same network. 

\textbf{Using depth in pose estimation} While using depth maps as additional data is largely unexplored in 3D pose estimation, it is popular in hand joint prediction \cite{wan2019selfsupervised-hand,cai2018weakhand,yuan2018privilegedhand}. In \cite{cai2018weakhand} the authors directly estimate the depth map from the predicted 3D coordinates. The estimation error is used as an additional regularizer.

Our work is closest to \cite{cai2018weakhand}. However, we do not predict a full depth map, rather just depth at individual points. This results in a smaller network and faster training.

\section{Method}

\begin{figure*}[t]
\begin{center}
\includegraphics[width=0.95\textwidth]{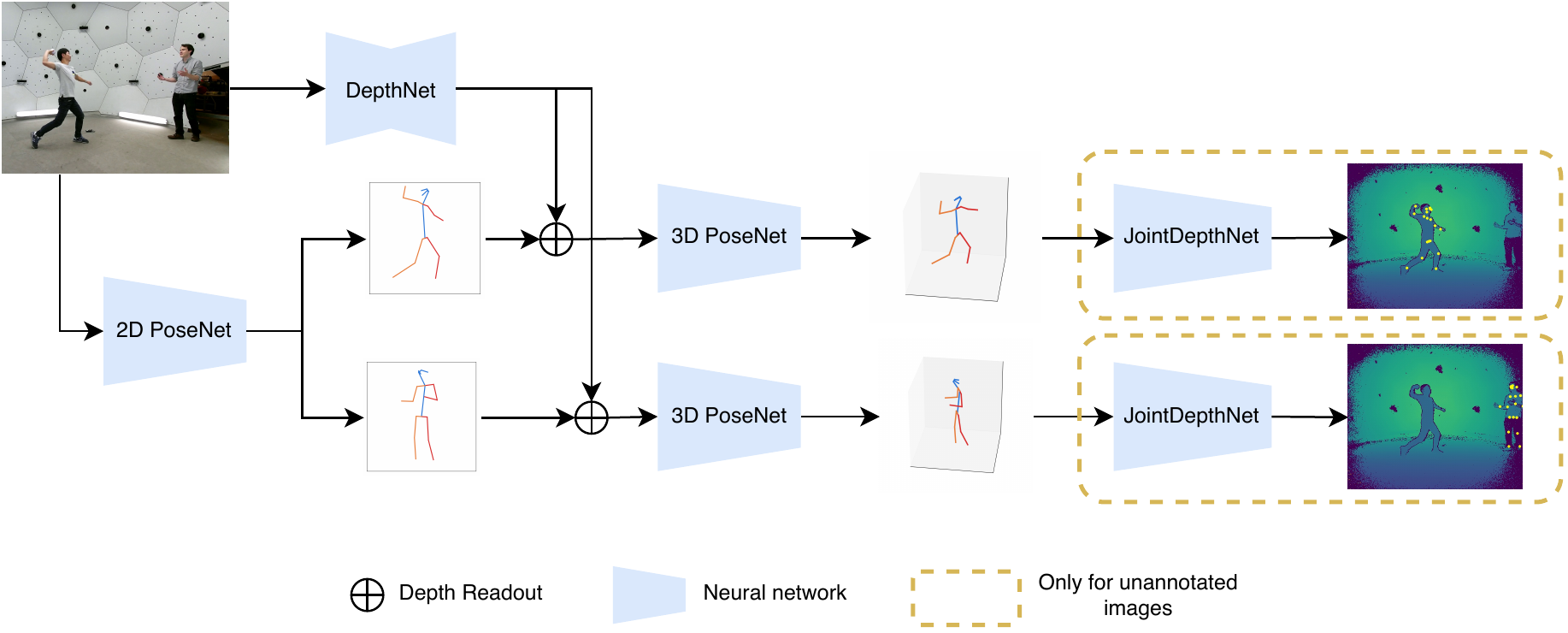}
\end{center}
\caption{\textbf{The network architecture.} The input image is fed to the 2D PoseNet and DepthNet. The 2D PoseNet detects humans on the picture and also returns the keypoint coordinates in pixel space. DepthNet estimates the depth for each pixel. Then the 2D pose and predicted depth are combined by reading out the predicted depth at the 2D joint locations. The 3D PoseNet predicts 3D poses from the concatenated 2D pose and depth features. The 3D estimation is performed for all poses separately. If the image does not have 3D pose annotation, the JointDepthNet estimates the depth at each joint on the ground-truth depth map (note that this is different from the depth of the joint because of occlusions).}
\label{fig:architecture}
\end{figure*}

The problem of absolute pose estimation can be formalized as follows: the algorithm must predict the $P^{3D}_j\in \mathbb{R}^3,1\le j\le J$ joint coordinates, where $J$ is the number of joints. The $P^{3D}_j$ points are in a camera-centered coordinate system where depth is perpendicular to the camera plane and the $x$ and $y$ axes align with the image edges.

\subsection{Overview} \label{sec:overview}
\begin{figure}[t]
\begin{center}
\begin{tabular}{ccc}
\includegraphics[width = 0.27\linewidth]{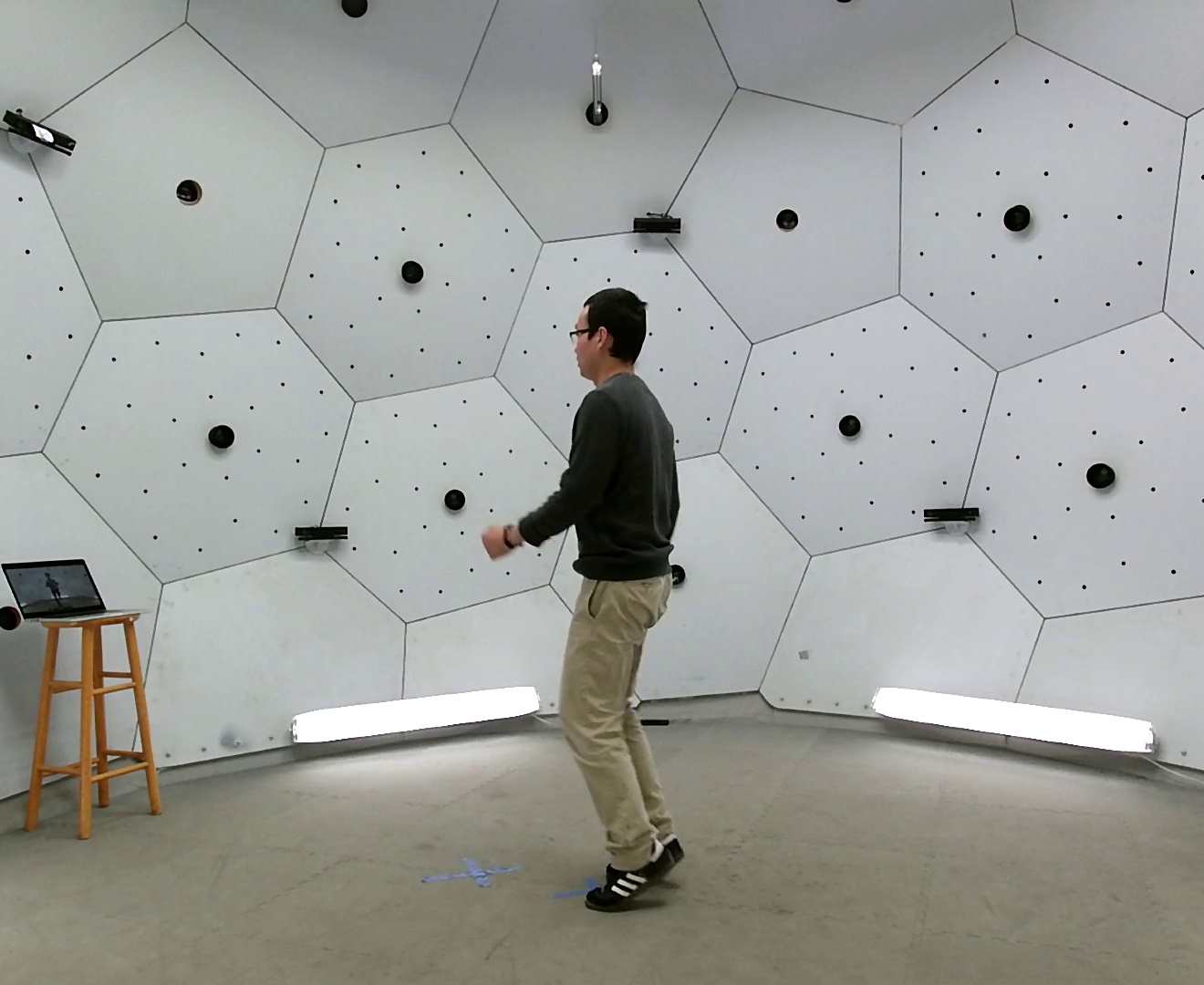} & 
\includegraphics[width = 0.27\linewidth]{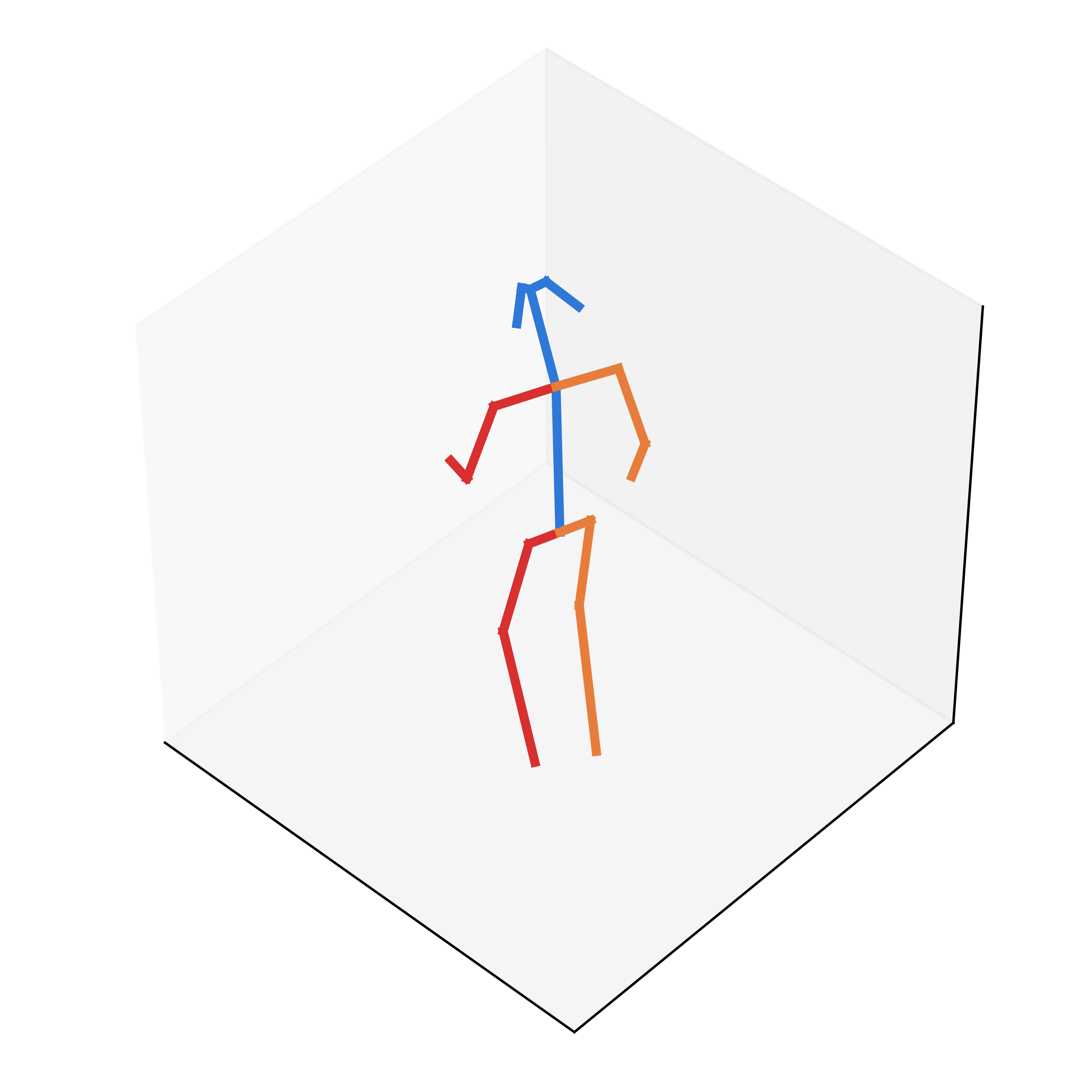} & 
\includegraphics[width = 0.27\linewidth]{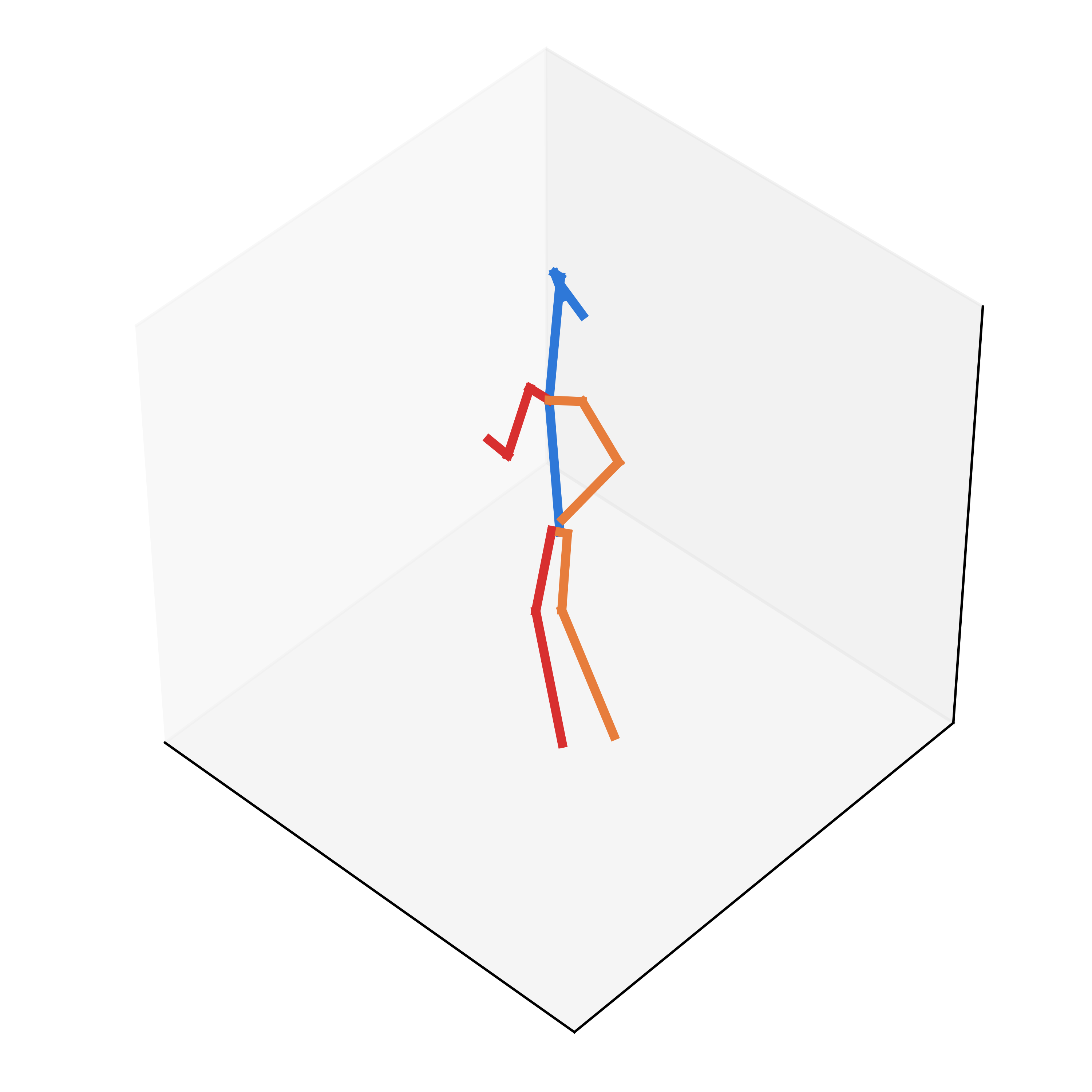}  \\
(a) & (b) & (c) \\
\end{tabular}
\end{center}
\caption{\textbf{Reconstructing a skeleton from the depth map under heavy self-occlusion.} a) Original image. b) Ground-truth 3D pose, from a different angle. c) Pose created by using ground-truth 2D coordinates and the depth from the depth images. Note how the (orange) right leg and arm are incorrect since the person is sideways on the image. Best viewed in color.}
\label{fig:self-occlusion}
\end{figure}

Our method is a multi-person, monocular estimator that takes single images as input. We use RGB\nobreakdash-D images as additional training examples, providing weak supervision. That is, we have two datasets: $\mathcal{D}=\left\{(I_1,P_1^{3D}), (I_2,P_2^{3D}),\ldots\right\}$, images with 3D pose annotations and $\mathcal{D}^*=\left\{(I_1,D_1), (I_2,D_2),\ldots\right\}$, images with pixelwise depth maps. On training images that have pose annotations, the prediction error can be calculated easily. For images from $\mathcal{D}^*$, one option would be to simply reconstruct the skeleton from the depth map and compare our prediction with that. However, due to self-occlusions and occlusion by other objects, the reconstructed skeleton would be noisy (see Figure \ref{fig:self-occlusion}). 

To solve this problem, we take a reverse approach: instead of comparing our predicted skeleton to a reconstructed one, we transform the predicted skeleton further, calculating the values of the depth map at the joint locations (see Fig.~\ref{fig:architecture} for the architecture). We introduce a network called \emph{JointDepthNet} that takes as input the predicted 3D pose $\hat{P}^{3D}$ and outputs  the value of the depth map at every joint $j$, denoted by $D^P_j$. It is important to note that $D^P_j$ is not the $z$ coordinate of the 3D pose in the camera coordinate system, but the depth detected by the depth sensor at the 2D position of joint $j$.  To get the 2D locations of $D^P_j$ we use the output of the 2D PoseNet (see Section \ref{sec:full-model}).

\subsection{Handling occlusions with robust loss} 
While JointDepthNet can resolve self-occlusions, it can not detect occlusion caused by the environment as it receives only the pose as input. 

We use the Geman--McClure robust loss function \cite{geman-mcclure} to overcome the large errors in the loss due to occlusions. The function is defined as:
$$\rho(x)=\frac{x^2}{x^2+\alpha},$$
where $\alpha$ is a parameter. The function is close to $x^2/\alpha$ around zero and converges to constant 1 for large numbers. 

Since JointDepthNet gets only the pose as input, it can not predict the depth at an occluded joint, resulting in a large error. If the error $x$ is large, then $\rho$ is approximately constant 1 around $x$, so the gradient is close to zero and such errors are effectively eliminated during training.

\subsection{Final model}\label{sec:full-model}

Our network is based on our previous work \cite{veges2019depthpose}, the architecture is illustrated in Fig \ref{fig:architecture}. We briefly review it here for completeness, please refer to \cite{veges2019depthpose} for details.

The network is a multi-stage architecture: the \emph{2D PoseNet} takes an image as input, and detects and predicts all the $P^{2D}_{(i)}$ 2D poses on the image, while \emph{DepthNet} predicts a single pixelwise depth map $D$. Then, for each pose $P^{2D}_{(i)}$, the depth at the joint locations are read out from $D$ and concatenated to the calibration matrix normalized $P^{2D}_{(i)}$. The normalization is needed by the model to handle different focal lengths and viewing angles. The concatenated vector is passed to the \emph{3D PoseNet} that predicts the 3D poses $P^{3D}_{(i)}$. If the input image does not have a pose annotation, then $P^{3D}_{(i)}$ is given to the JointDepthNet that resolves occlusions and outputs the $D^P_{j,(i)},\, 1\le j\le J$ depth values (see Section \ref{sec:overview}). 

JointDepthNet is run only for unannotated images during training. In inference and for images with pose annotations it is ignored.

For pose annotated images, the loss is the L1 loss, for unannotated images it is the Geman--McClure loss. The total loss is then:
$$L=\sum_{I\in \mathcal{D}}L_1\left(\hat{P}^{3D}_{(i)},P^{3D}_{(i)}\right)+\lambda\sum_{I\in \mathcal{D}^*}\sum_{j=1}^J\rho\left(\hat{D}^P_{j,(i)}-D^P_{j,(i)}\right),$$
where $\lambda$ is a weight, $\hat{P}^{3D}_{(i)}$ are the predicted poses, and $\hat{D}^P_{j,(i)}$ are the predicted depths at each joint $j$.

One difference compared to \cite{veges2019depthpose} is that instead of Batch Normalization layers \cite{batchnorm} we used Layer Normalization \cite{ba2016layer}. We have found that normalizing over the entire batch while mixing two datasets led to suboptimal results, since the characteristics of the two databases were different. When normalizing each training example separately, the performance of the network increased, see Table \ref{tbl:ablation}.

\section{Experiments}
\subsection{Datasets} \label{sec:datasets}

\textbf{MuPoTS-3D} We conducted experiments on the MuPoTS-3D database \cite{mehta2018single_shot}. It is a multi-person dataset having both indoor and outdoor videos. In the standard protocol, the training set is the MuCo-3DHP dataset \cite{mehta2018single_shot} that contains synthetic images, composited from frames in the MPI-INF-3DHP \cite{mehta} database. We have created 150k images, each containing 4 persons. The training and test set are quite different so it is a good measurement of in-the-wild robustness.

In addition to the raw annotations, the MuCo-3DHP and MuPoTS-3D databases contains normalized skeletons. The normalization process rescales the skeleton from the hip such that the knee to neck distance becomes a fixed value. Prior work uses either one or the other skeleton, we evaluate our method on both annotations for completeness. Note that using unnormalized skeletons for absolute pose estimation is more principled. Normalized poses were proposed for the relative pose estimation task where only the orientation and angles of limbs are relevant. In contrast, in absolute pose estimation the location of the joints is important too. Applying a hip-centered scaling on the skeleton leaves the hip in the correct position, while all the other joints are moved to an incorrect location.

\textbf{Panoptic} We also performed experiments on the Panoptic dataset \cite{panoptic}. The dataset consists of multiple RGB\nobreakdash-D videos, recorded by Kinect sensors, from multiple viewpoints.

Since there is no standard training/test set defined, we selected one session as test and another as validation. The test/validation split contained the recordings of all the RGB-D cameras for the selected session.

\subsection{Evaluation metrics}
\begin{table*}[h]
\caption{\textbf{Relative-3DPCK on the MuPoTS-3D dataset (normalized skeletons).} Comparison with previous work that uses normalized coordinates (see text for details). Each column corresponds to a video sequence. Higher values are better. Our results are competitive to the state-of-the-art.}
\begin{center}
\resizebox{0.95\textwidth}{!}{
\begin{tabular}{lrrrrrrrrrrrrrrrrrrrr||r}
\hline
Method & S1 & S2 & S3 & S4 & S5 & S6 & S7 & S8 & S9 & S10 & S11 & S12 & S13 & S14 & S15 & S16 & S17 & S18 & S19 & S20 & Avg. \\
\hline
\multicolumn{21}{l||}{\emph{Comparing every pose}} &  \\
Rogez \cite{rogez2017lcrnet} & 67.7 & 49.8 & 53.4 & 59.1 & 67.5 & 22.8 & 43.7 & 49.9 & 31.1 & 78.1 & 50.2 & 51.0 & 51.6 & 49.3 & 56.2 & 66.5 & 65.2 & 62.9 & 66.1 & 59.1 & 53.8\\
Mehta \cite{mehta2018single_shot} & 81.0 & 60.9 & 64.4 & 63.0 & 69.1 & 30.3 & 65.0 & 59.6 & 64.1 & 83.9 & 68.0 & 68.6 & 62.3 & 59.2 & 70.1 & 80.0 & 79.6 & 67.3 & 66.6 & 67.2 & 66.0\\
Rogez \cite{rogez2019lcrpp} & 87.3 & 61.9 & 67.9 & 74.6 & 78.8 & 48.9 & 58.3 & 59.7 & 78.1 & 89.5 & 69.2 & 73.8 & 66.2 & 56.0 & 74.1 & 82.1 & 78.1 & 72.6 & 73.1 & 61.0 & 70.6\\
Moon \cite{moon2019camdistance} & \textbf{94.4} & \textbf{77.5} & 79.0 & 81.9 & \textbf{85.3} & 72.8 & 81.9 & \textbf{75.7} & \textbf{90.2} & 90.4 & \textbf{79.2} & 79.9 & \textbf{75.1} & \textbf{72.7} & \textbf{81.1} & 89.9 & 89.6 & \textbf{81.8} & \textbf{81.7} & \textbf{76.2} & \textbf{81.8}\\
\textbf{Ours} & 89.5 & 75.9 & \textbf{85.2} & \textbf{83.9} & 85.0 & \textbf{73.4} & \textbf{83.6} & 58.7 & 65.1 & \textbf{90.4} & 76.8 & \textbf{81.9} & 67.0 & 55.9 & 80.8 & \textbf{90.6} & \textbf{90.0} & 81.1 & 81.1 & 68.6 & 78.2\\
\hline
\multicolumn{21}{l||}{\emph{Comparing detected poses only}} \\
Rogez \cite{rogez2017lcrnet} & 69.1 & 67.3 & 54.6 & 61.7 & 74.5 & 25.2 & 48.4 & 63.3 & 69.0 & 78.1 & 53.8 & 52.2 & 60.5 & 60.9 & 59.1 & 70.5 & 76.0 & 70.0 & 77.1 & 81.4 & 62.4\\
Mehta \cite{mehta2018single_shot} & 81.0 & 65.3 & 64.6 & 63.9 & 75.0 & 30.3 & 65.1 & 61.1 & 64.1 & 83.9 & 72.4 & 69.9 & 71.0 & 72.9 & 71.3 & 83.6 & 79.6 & 73.5 & 78.9 & 90.9 & 70.8\\
Rogez \cite{rogez2019lcrpp} & 88.0 & 73.3 & 67.9 & 74.6 & 81.8 & 50.1 & 60.6 & 60.8 & 78.2 & 89.5 & 70.8 & 74.4 & 72.8 & 64.5 & 74.2 & 84.9 & 85.2 & 78.4 & 75.8 & 74.4 & 74.0\\
Moon \cite{moon2019camdistance} & \textbf{94.4} & 78.6 & 79.0 & 82.1 & 86.6 & 72.8 & 81.9 & \textbf{75.8} & \textbf{90.2} & 90.4 & \textbf{79.4} & 79.9 & \textbf{75.3} & \textbf{81.0} & 81.0 & 90.7 & 89.6 & 83.1 & 81.7 & 77.3 & 82.5\\
\textbf{Ours} & 89.5 & \textbf{81.6} & \textbf{85.9} & \textbf{84.4} & \textbf{90.5} & \textbf{73.5} & \textbf{85.5} & 68.9 & 65.1 & \textbf{90.4} & 79.1 & \textbf{82.6} & 72.7 & 68.1 & \textbf{81.0} & \textbf{94.0} & \textbf{90.4} & \textbf{87.4} & \textbf{90.4} & \textbf{92.6} & \textbf{82.7}\\
\hline
\end{tabular}}
\end{center}
\label{tbl:mupots-rel}
\end{table*}

\begin{table*}[h]
\caption{\textbf{Absolute-3DPCK on the MuPoTS-3D dataset (normalized skeletons).} Comparison with previous work that uses normalized coordinates (see text for details). Each column corresponds to a video sequence. Higher values are better. We achieve state-of-the-art results both for detected poses and all poses.}
\begin{center}
\resizebox{0.95\textwidth}{!}{
\begin{tabular}{lrrrrrrrrrrrrrrrrrrrr||r}
\hline
Method & S1 & S2 & S3 & S4 & S5 & S6 & S7 & S8 & S9 & S10 & S11 & S12 & S13 & S14 & S15 & S16 & S17 & S18 & S19 & S20 & Avg. \\
\hline
\multicolumn{21}{l||}{\emph{Comparing every pose}} & \\
Moon \cite{moon2019camdistance} & \textbf{59.5} & \textbf{44.7} & 51.4 & \textbf{46.0} & 52.2 & 27.4 & \textbf{23.7} & 26.4 & \textbf{39.1} & 23.6 & 18.3 & 14.9 & 38.2 & 26.5 & 36.8 & \textbf{23.4} & 14.4 & 19.7 & 18.8 & 25.1 & 31.5\\
\textbf{Ours} & 50.4 & 33.4 & \textbf{52.8} & 27.5 & \textbf{53.7} & \textbf{31.4} & 22.6 & \textbf{33.5} & 38.3 & \textbf{56.5} & \textbf{24.4} & \textbf{35.5} & \textbf{45.5} & \textbf{34.9} & \textbf{49.3} & 23.2 & \textbf{32.0} & \textbf{30.7} & \textbf{26.3} & \textbf{43.8} & \textbf{37.3}\\
\hline
\multicolumn{21}{l||}{\emph{Comparing detected poses only}} & \\
Moon \cite{moon2019camdistance} & \textbf{59.5} & \textbf{45.3} & 51.4 & \textbf{46.2} & 53.0 & 27.4 & \textbf{23.7} & 26.4 & \textbf{39.1} & 23.6 & 18.3 & 14.9 & 38.2 & 29.5 & 36.8 & 23.6 & 14.4 & 20.0 & 18.8 & 25.4 & 31.8\\
\textbf{Ours} & 50.4 & 35.9 & \textbf{53.3} & 27.7 & \textbf{57.2} & \textbf{31.4} & 23.1 & \textbf{39.3} & 38.3 & \textbf{56.5} & \textbf{25.2} & \textbf{35.8} & \textbf{49.3} & \textbf{42.5} & \textbf{49.4} & \textbf{24.1} & \textbf{32.1} & \textbf{33.1} & \textbf{29.3} & \textbf{59.2} & \textbf{39.6}\\
\hline
\end{tabular}}
\end{center}
\label{tbl:mupots-abs}
\end{table*}

Different works use different evaluation metrics for absolute pose estimation. For completeness we evaluate our method on all of them. We shortly review these below:

\textbf{A-MPJPE} or Absolute Mean per Joint Position Error \cite{veges2019depthpose}. It is the Euclidean distance between the ground-truth and predicted joints, averaged over all poses and joints. The metric has the drawback that it does not take into account undetected poses. We present it in mm in our results.

\textbf{R-MPJPE} or Relative Mean per Joint Position Error \cite{h36m,veges2019depthpose}. It is the same as A-MPJPE but with root joints moved to the ground-truth location. The metric is often referred as MPJPE in other work. Presented in mm everywhere.

\textbf{A-3DPCK} or Absolute 3D Percentage of Correct Keypoints \cite{moon2019camdistance}. It is the percentage of keypoints where the prediction error is less then 15~cm. If a pose is not detected, then the prediction error is defined as infinite thus it does not contribute to the metric. In contrast to A-MPJPE, the metric is sensitive to undetected poses.

\textbf{R-3DPCK} or Relative 3D Percentage of Correct Keypoints \cite{mehta2018single_shot,moon2019camdistance}, commonly referred as 3DPCK. Same as A\nobreakdash-3DPCK but with root joints moved to the ground-truth position. It is the standard metric for relative pose estimation on the MuPo-TS dataset. Similarly to A\nobreakdash-3DPCK, it takes into account undetected poses.

\subsection{Implementation details}
Our network architecture is based on \cite{veges2019depthpose}. For the 2D PoseNet we selected the state-of-the-art HR-Net pose estimator \cite{hrnet} with Mask-RCNN \cite{maskrcnn} as the human bounding-box detector; for DepthNet, following \cite{veges2019depthpose}, we used MegaDepth \cite{megadepth2018}. The 3D PoseNet consists of two residual blocks, each having two fully connected layers of 1024 neurons. The dense layers are followed by a Layer Normalization, Dropout and ReLU activation layers. The JointDepthNet has the same structure, having two residual blocks. The dropout rate was 0.5. We normalized the poses to have standard deviation of 1 and zero mean before training and split it to relative pose and root joint localization. See \cite{veges2019depthpose} for details. JointDepthNet predicts $D^P_i$ for only 14 joints  that we found to be stable, these are the wrists, elbows, shoulders, hips, knees and elbows.

The 3D PoseNet and JointDepthNet were trained jointly, using Adam with a learning rate of 0.001. Every four epoch, the learning rate was multiplied by a factor of 0.96. Half of a mini-batch contained images with pose annotations and the other half contained images with depth map only.
The network was trained for 100 epochs.

During training, we applied image augmentation by randomly zooming into the images, while camera intrinsics remained unchanged. This augmentation was performed both for images in $\mathcal{D}$ and $\mathcal{D}^*$, while the target depth maps and poses were appropriately scaled. With this setup, zooming corresponds to moving the poses closer or further away from the camera. We found that this step is essential, otherwise the network overfits to the $y$ locations in the training set.

\section{Results}
\subsection{Quantitative results}

We evaluated our model quantitatively on the MuPoTS-3D and Panoptic datasets. 

On MuPoTS-3D, we trained our model on the MuCo and on Panoptic datasets jointly, using only depth maps from the latter (\emph{Ours} in the results). Previous work either used the raw unnormalized, or height-normalized coordinates (see Section \ref{sec:datasets} for details). 

The absolute pose estimation results on normalized coordinates are shown in Table \ref{tbl:mupots-abs}. The A\nobreakdash-3DPCK metric can be calculated on all poses or only on the detected ones. The latter is useful to asses the pose estimation performance, while the former also takes into account the detection performance. On all poses, our model achieved 37.3\%, which is 5.8\% larger than the previous state of the art. This corresponds to a 18.4\% relative increase in the metric. On detected poses only, we improved  A\nobreakdash-3DPCK by 7.8\% (24.5\% relatively). Our model remains competitive with the state-of-the-art in the relative pose prediction metrics (Table \ref{tbl:mupots-rel}).

Table \ref{tbl:mupots-normal} compares our method to prior work using unnormalized coordinates on MuPoTS-3D. The authors of \cite{veges2019depthpose} evaluated their method using the MPJPE metrics. Our method decreases the A-MPJPE and R-MPJPE error by 37~mm and 12~mm (12.6\% and 10\% relatively). Moreover, our model's detection rate is also higher (93\% vs 91\%).

\begin{table}[h]
\caption{\textbf{Results on the MuPoTS-3D dataset (unnormalized skeletons).} Comparison with previous work that uses unnormalized coordinates (see text for details). Metrics are calculated on detected poses only. MPJPE errors are in mm. }\label{tbl:mupots-normal}
\begin{center}
\begin{tabular}{lccccc}
 & A-MPJPE $\downarrow$ & R-MPJPE $\downarrow$ &  A-3DPCK $\uparrow$  & R-3DPCK $\uparrow$ & Det. Rate $\uparrow$   \\
\hline
LCR-Net \cite{rogez2017lcrnet} & - & 146  & - & - & 86\% \\
Mehta et al. \cite{mehta2018single_shot} & - & 132  & - & - & 93\% \\ 
Veges et al. \cite{veges2019depthpose} & 292 & 120  & 30.1 & - & 91\% \\
\hline
\textbf{Ours} & \textbf{255} & \textbf{108}  & \textbf{35.9} & \textbf{78.7} & \textbf{93\%} \\
\hline
\end{tabular}
\end{center}
\end{table}

\begin{table}[h]
\caption{\textbf{Results on the Panoptic dataset.} \emph{Ours w/o JDN} is our network without JointDepthNet, using only frames with pose annotations. Evaluating on detected poses only. MPJPE errors are in mm. Ties are not marked in bold. To our knowledge, no other work predicts absolute poses on Panoptic. }
\label{tbl:panoptic}
\begin{center}
\begin{tabular}{lccccc}
\hline
 & Fine-tuned & A-MPJPE $\downarrow$ & A-3DPCK $\uparrow$ & R-MPJPE $\downarrow$ & R-3DPCK $\uparrow$   \\
\hline
Ours w/o JDN & No & 151 & 60.6  & \textbf{67.4} & 95.2  \\  
Ours & No & \textbf{147} & \textbf{62.5} & 67.6 & 95.2  \\ 
\hline
Ours w/o JDN & Yes & 144 & 64.1 & \textbf{62.2} & \textbf{96.6}  \\  
Ours & Yes & \textbf{134} & \textbf{68.8}  & 62.3 & 96.5   \\ 
\hline
\end{tabular}
\end{center}
\end{table}

Finally, we evaluate our algorithm on the Panoptic dataset. The database contains both depth maps and 3D pose annotations, thus we split the training set in two parts, one part uses only pose annotations, the other only depth maps. The results are shown in Table \ref{tbl:panoptic}. The weak supervision improves both A\nobreakdash-MPJPE and A\nobreakdash-3DPCK, while the relative metrics remain unchanged or change slightly.

The 3D pose estimator has a depth estimator component. Since we have depth images in the Panoptic dataset, it is natural to investigate, whether fine-tuning the DepthNet on images from $\mathcal{D}^*$ negates the improvement from JointDepthNet. We performed this experiment (Fine-tuned: Yes in Table \ref{tbl:panoptic}). Results show that the weak supervision still improves the performance, in fact with larger margin than without fine-tuning.

\subsection{Qualitative results}
\begin{figure*}
\begin{center}
\begin{tabular}{cccccc}
\begin{minipage}[c]{0.15\linewidth}\includegraphics[width = \textwidth]{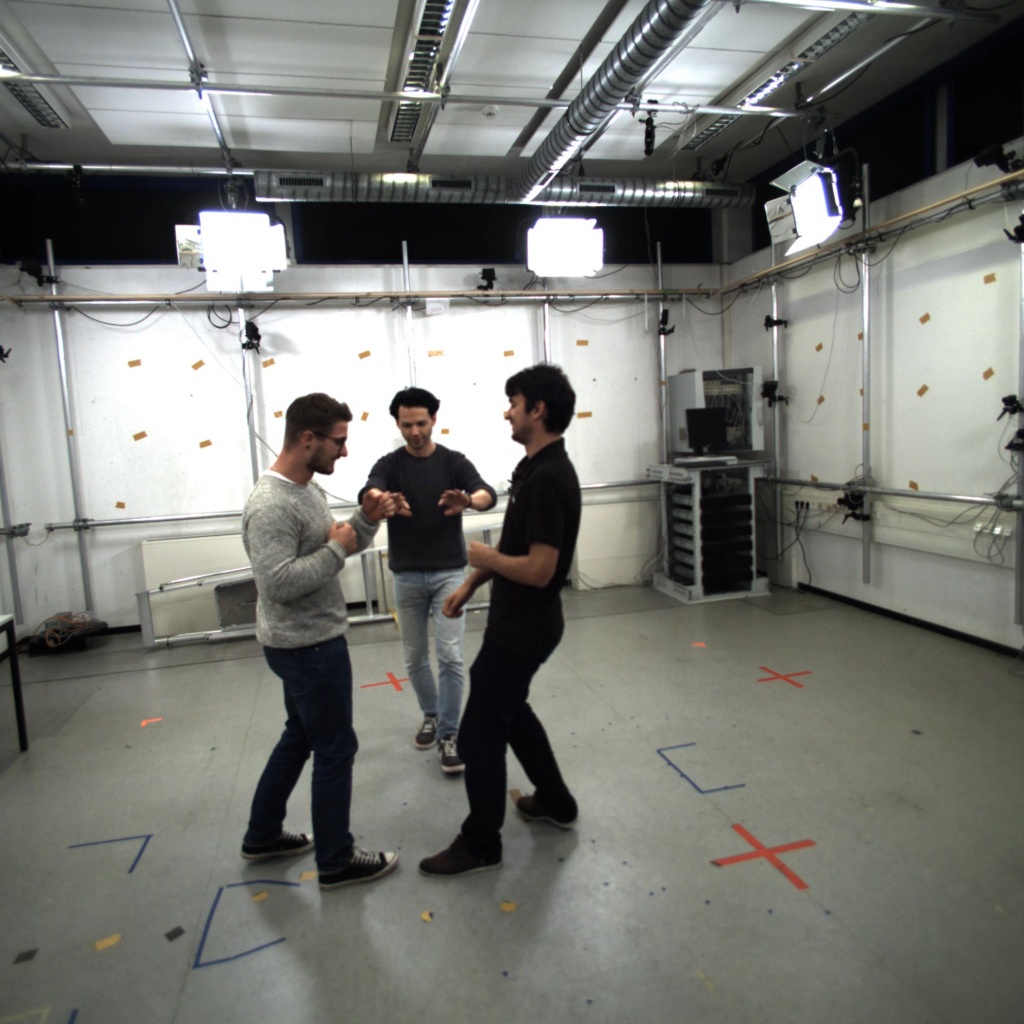}\end{minipage} & 
\begin{minipage}[c]{0.15\linewidth}\includegraphics[width = \textwidth]{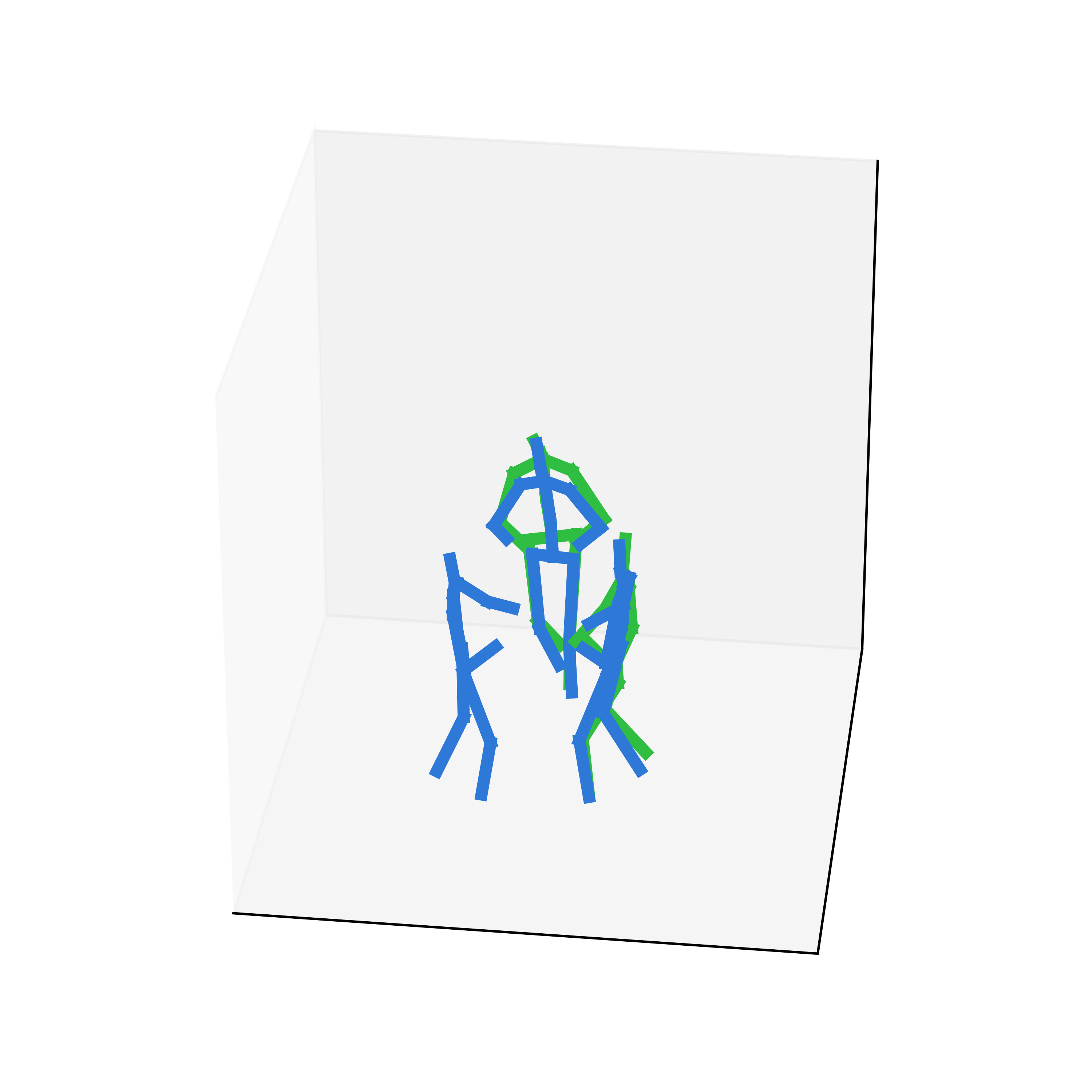} \end{minipage} &
\begin{minipage}[c]{0.15\linewidth}\includegraphics[width = \textwidth]{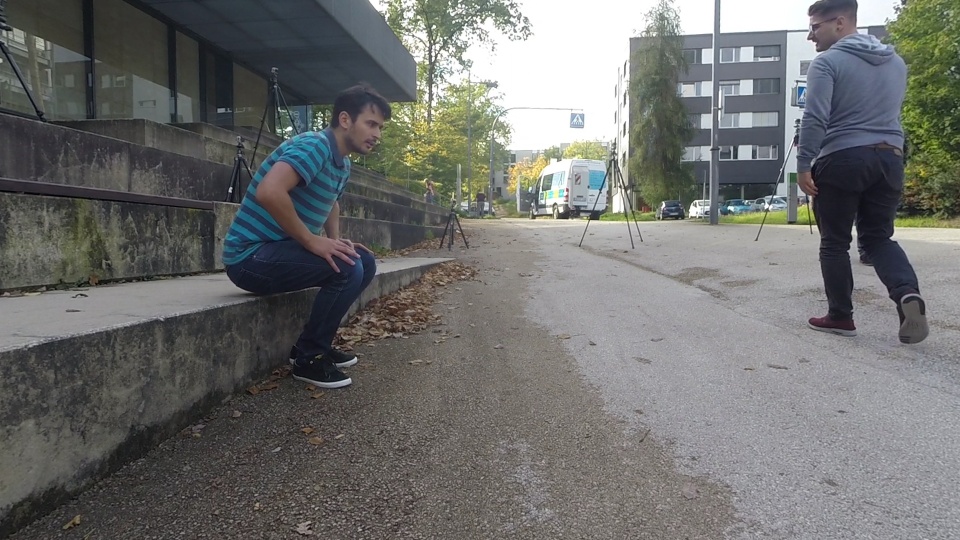}\end{minipage} & 
\begin{minipage}[c]{0.15\linewidth}\includegraphics[width = \textwidth]{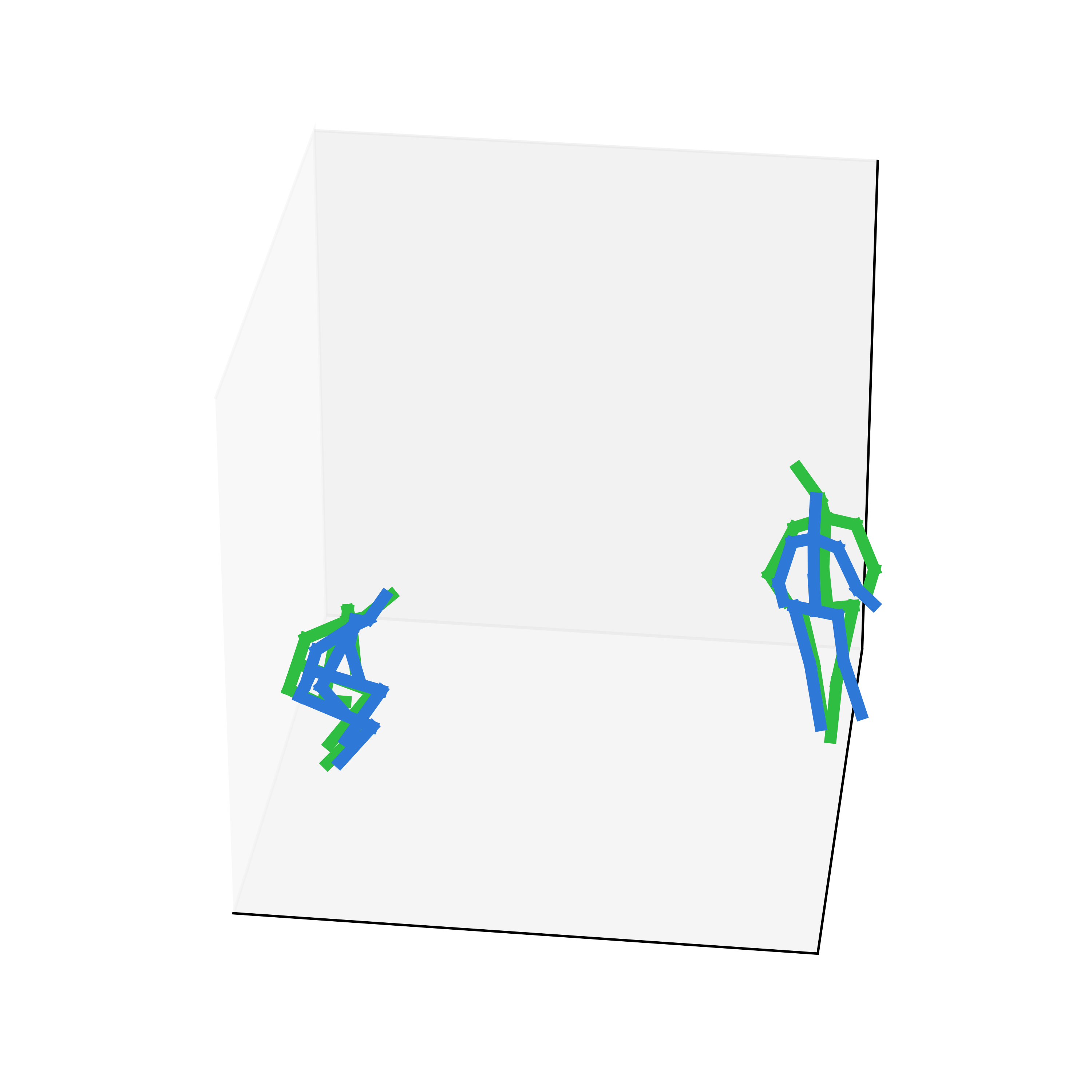} \end{minipage} &
\begin{minipage}[c]{0.15\linewidth}\includegraphics[width = \textwidth]{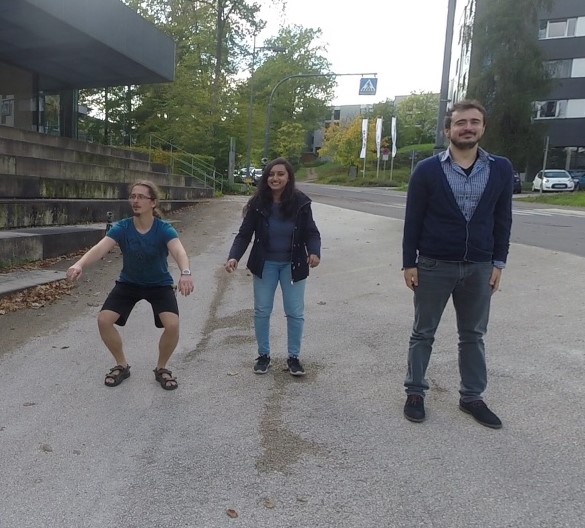}\end{minipage} & 
\begin{minipage}[c]{0.15\linewidth}\includegraphics[width = \textwidth]{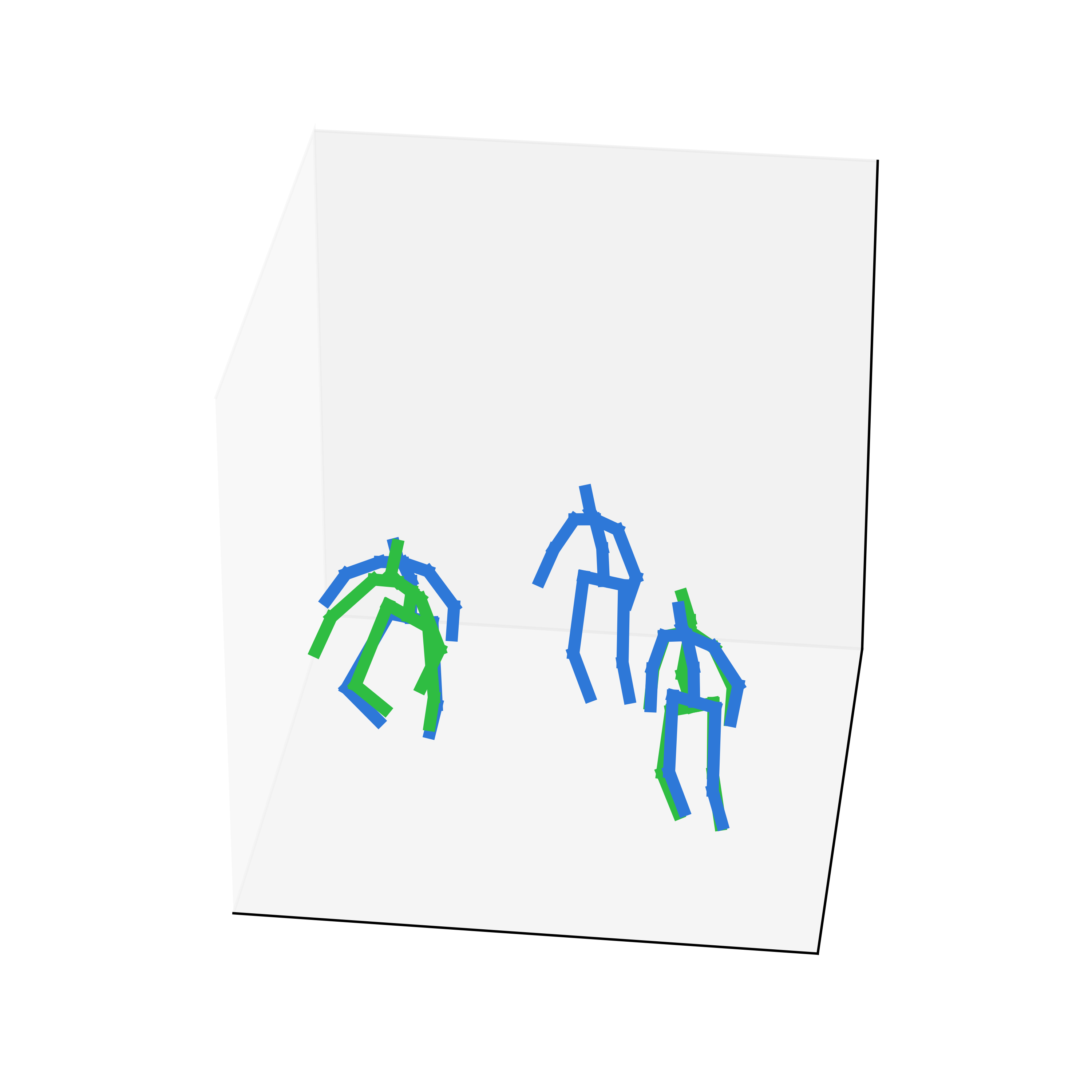} \end{minipage} \\

\begin{minipage}[c]{0.15\linewidth}\includegraphics[width = \textwidth]{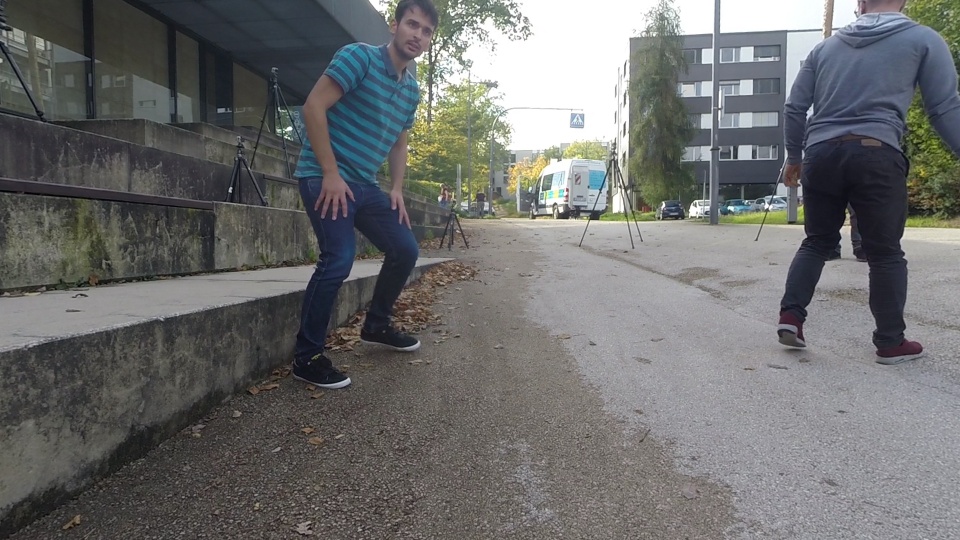}\end{minipage} & 
\begin{minipage}[c]{0.15\linewidth}\includegraphics[width = \textwidth]{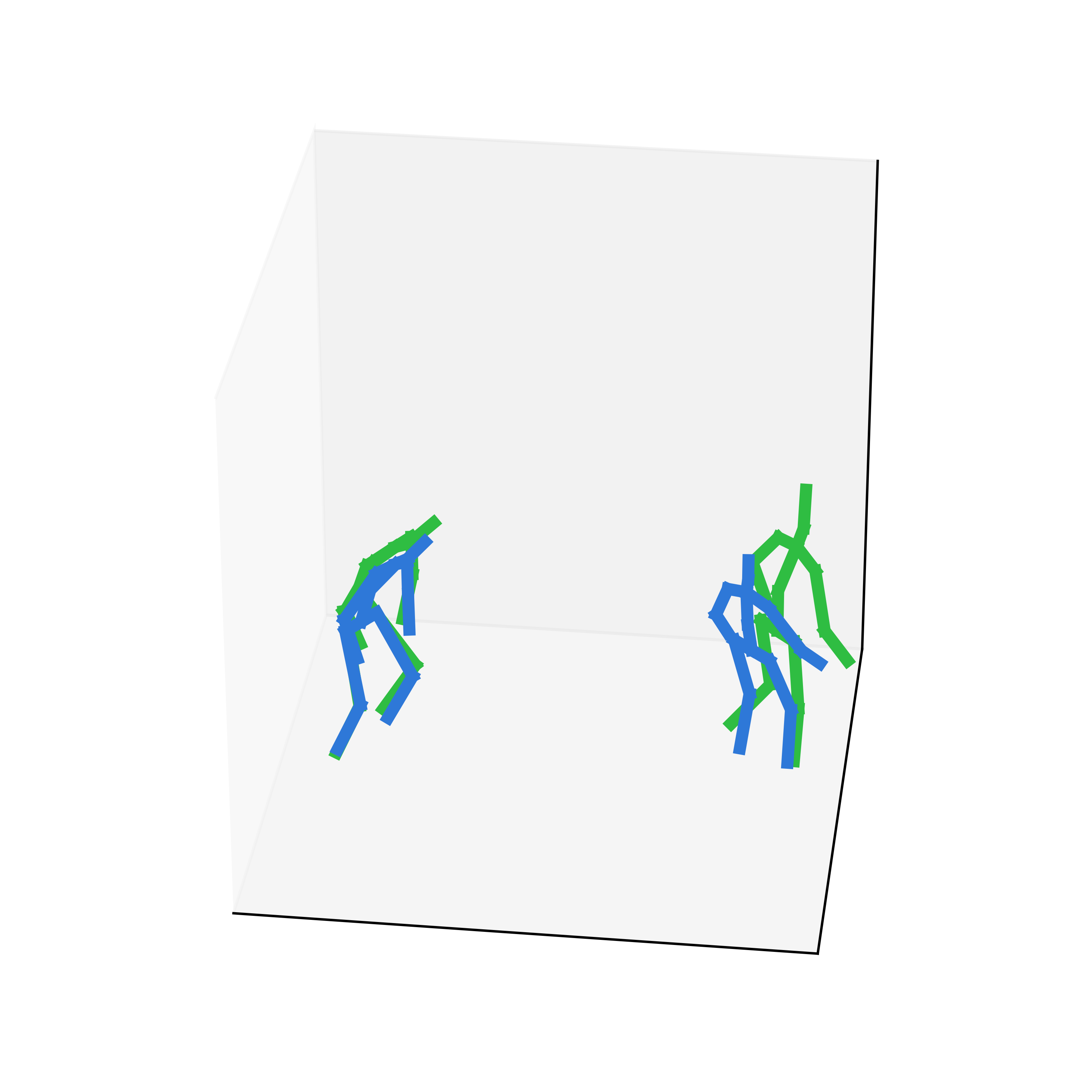} \end{minipage} &
\begin{minipage}[c]{0.15\linewidth}\includegraphics[width = \textwidth]{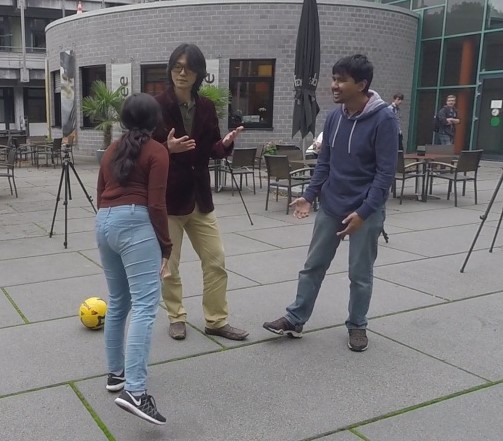}\end{minipage} & 
\begin{minipage}[c]{0.15\linewidth}\includegraphics[width = \textwidth]{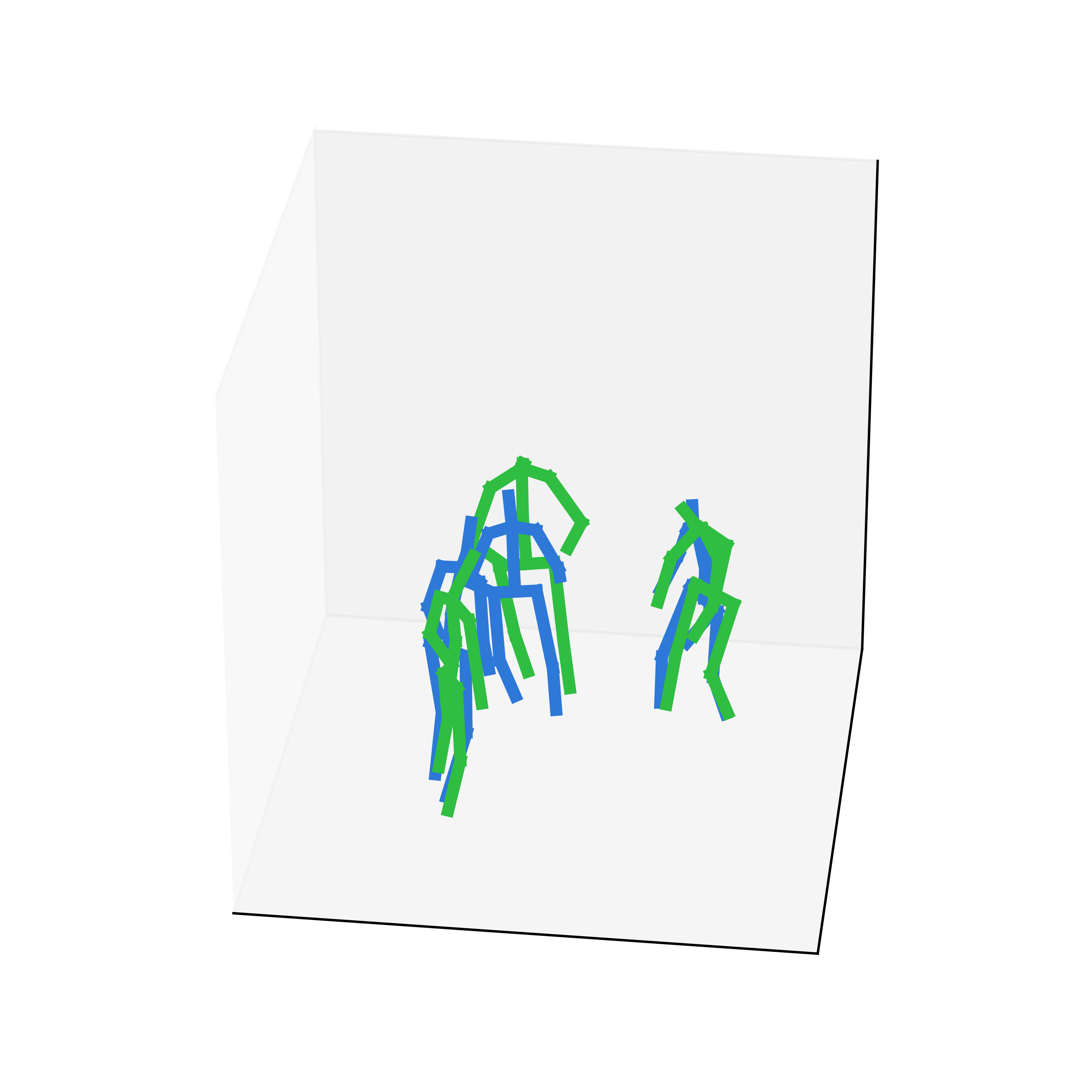} \end{minipage} &
\begin{minipage}[c]{0.15\linewidth}\includegraphics[width = \textwidth]{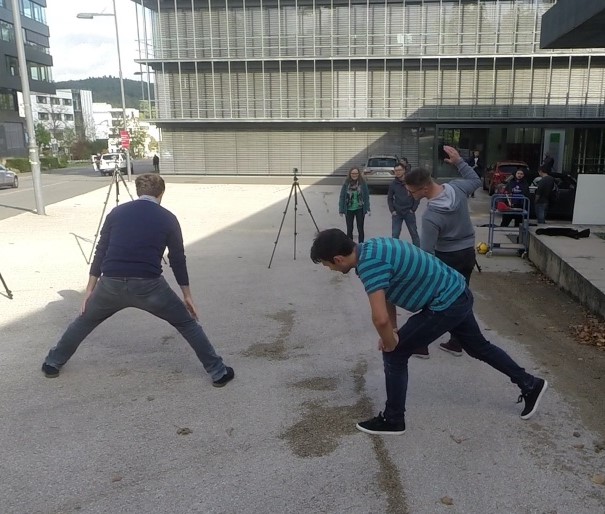}\end{minipage} & 
\begin{minipage}[c]{0.15\linewidth}\includegraphics[width = \textwidth]{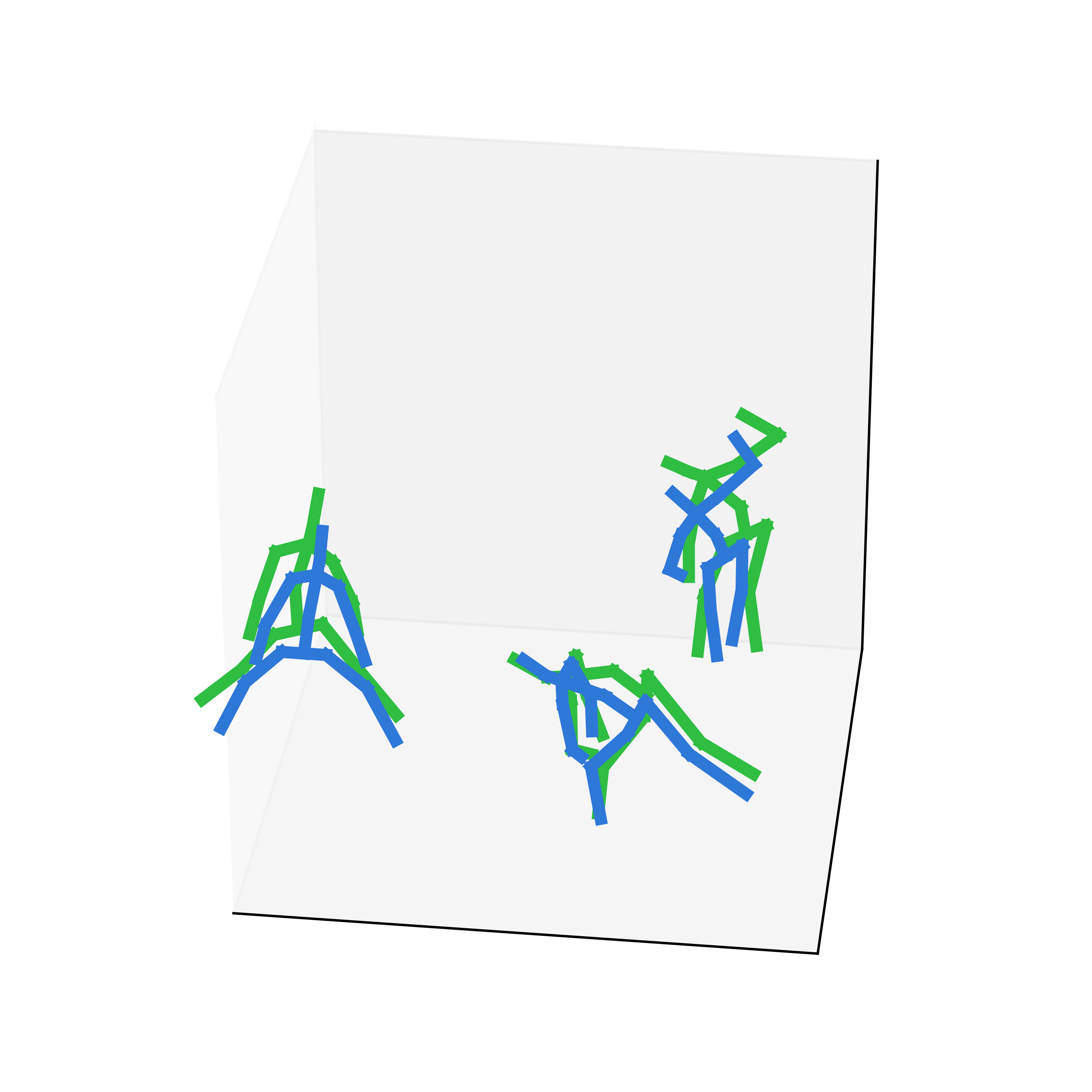} \end{minipage} \\

\begin{minipage}[c]{0.15\linewidth}\includegraphics[width = \textwidth]{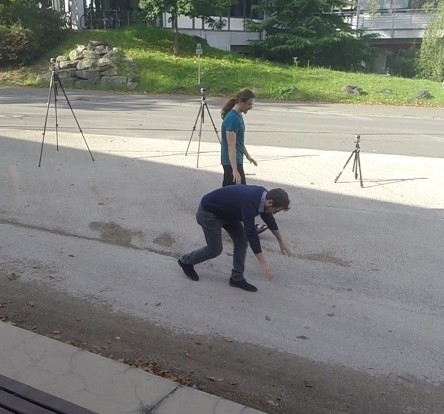}\end{minipage} & 
\begin{minipage}[c]{0.15\linewidth}\includegraphics[width = \textwidth]{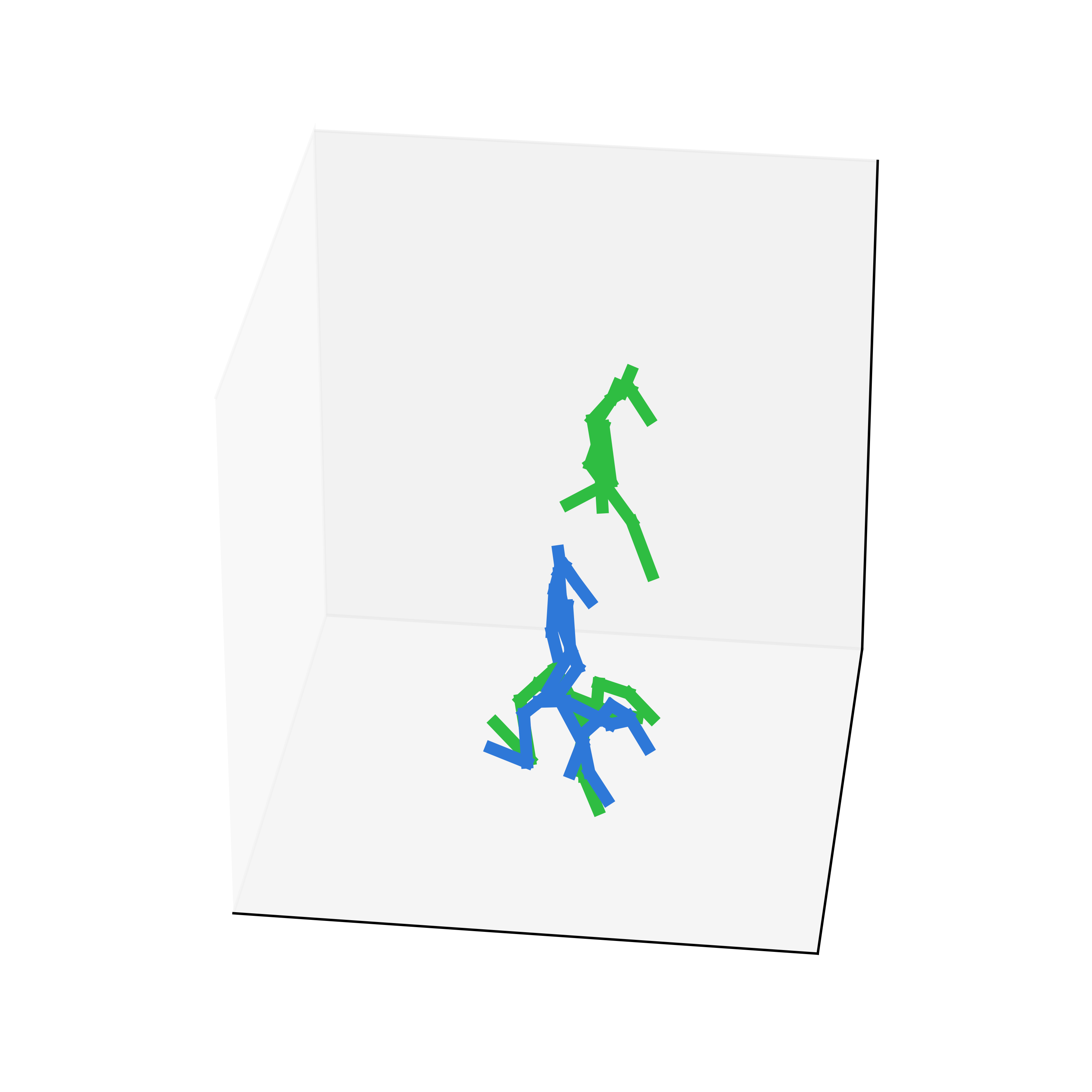} \end{minipage} &
\begin{minipage}[c]{0.15\linewidth}\includegraphics[width = \textwidth]{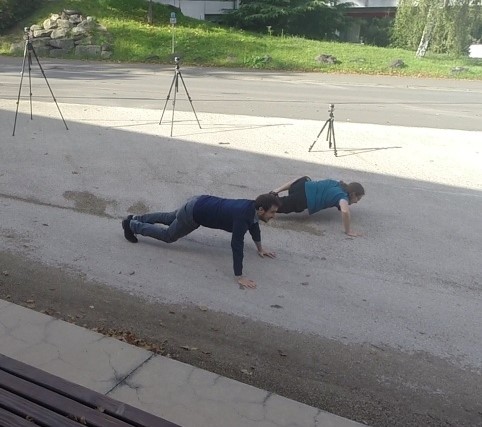}\end{minipage} & 
\begin{minipage}[c]{0.15\linewidth}\includegraphics[width = \textwidth]{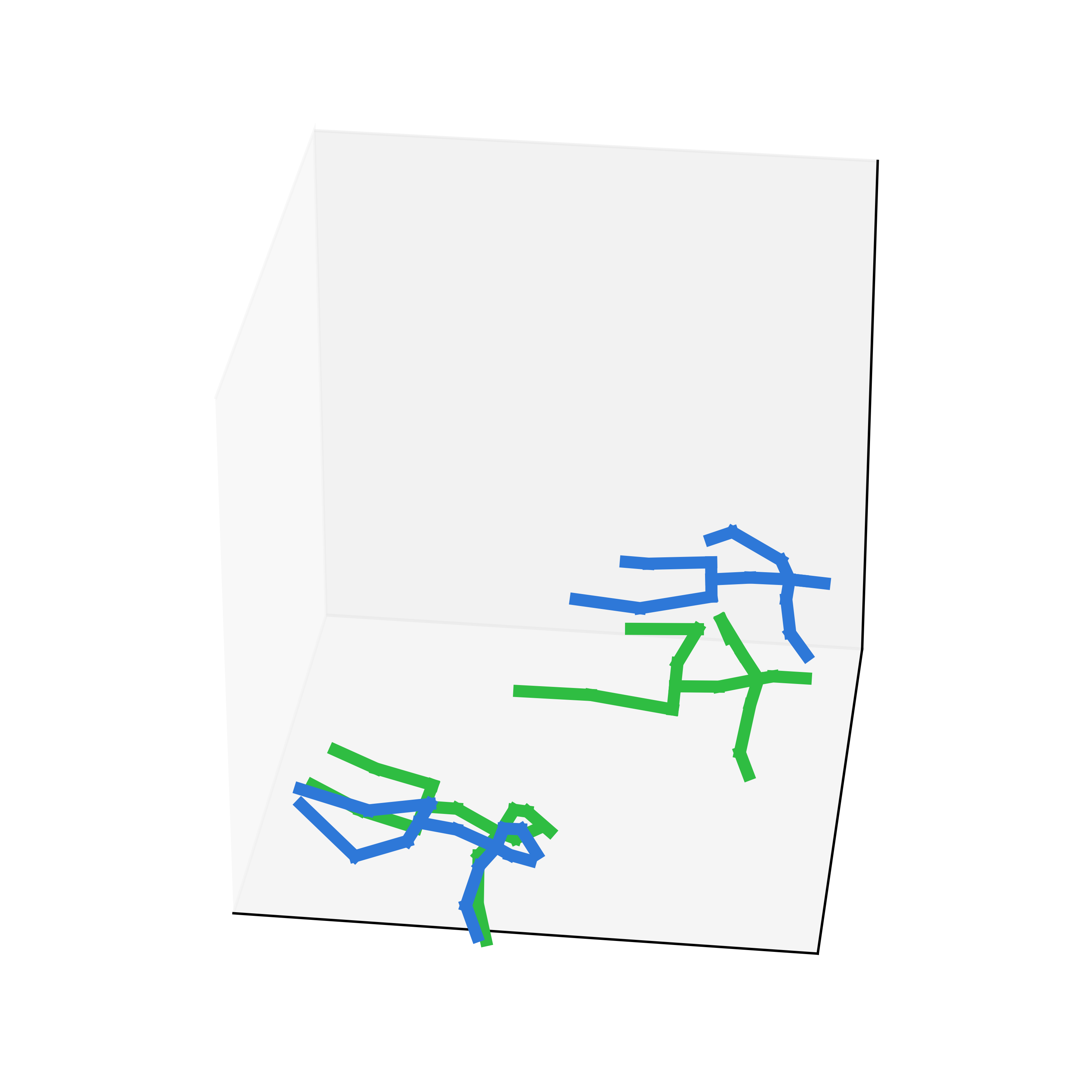} \end{minipage} &
\begin{minipage}[c]{0.15\linewidth}\includegraphics[width = \textwidth]{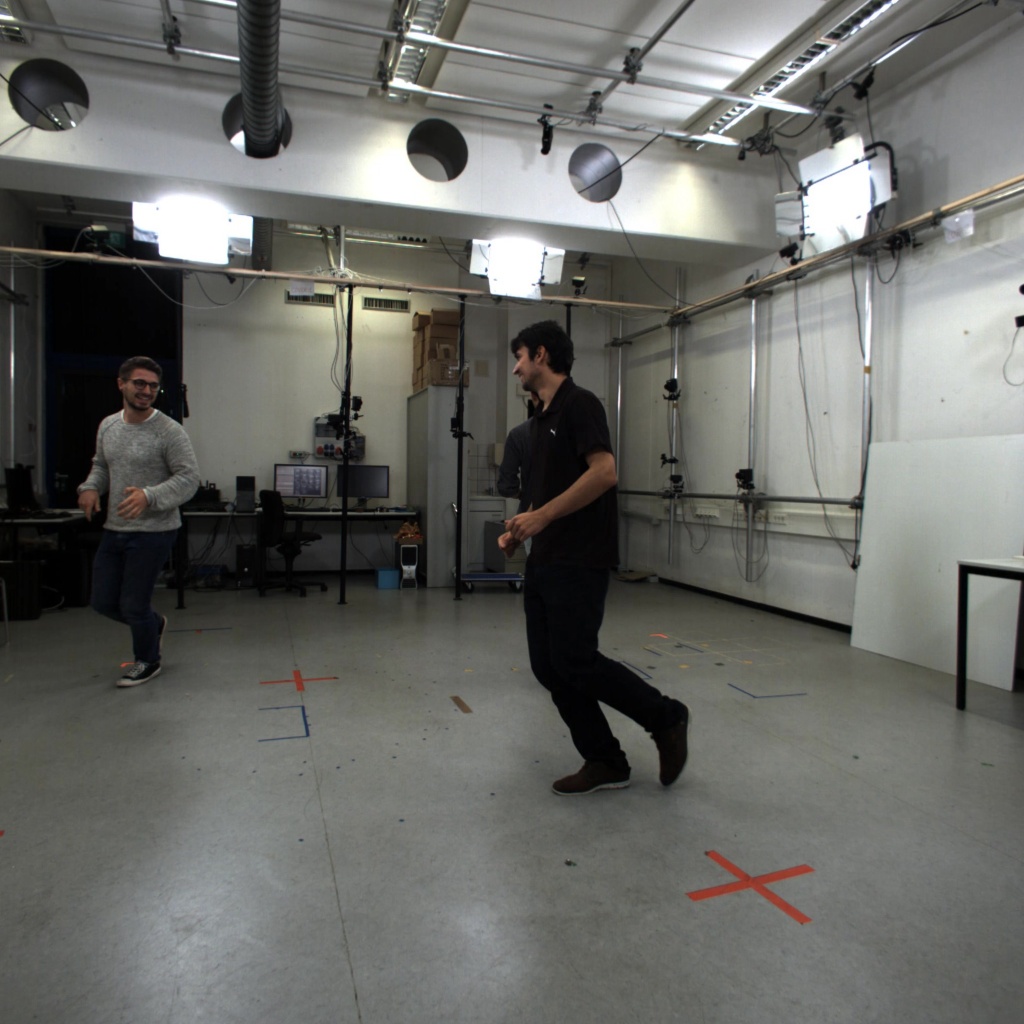}\end{minipage} & 
\begin{minipage}[c]{0.15\linewidth}\includegraphics[width = \textwidth]{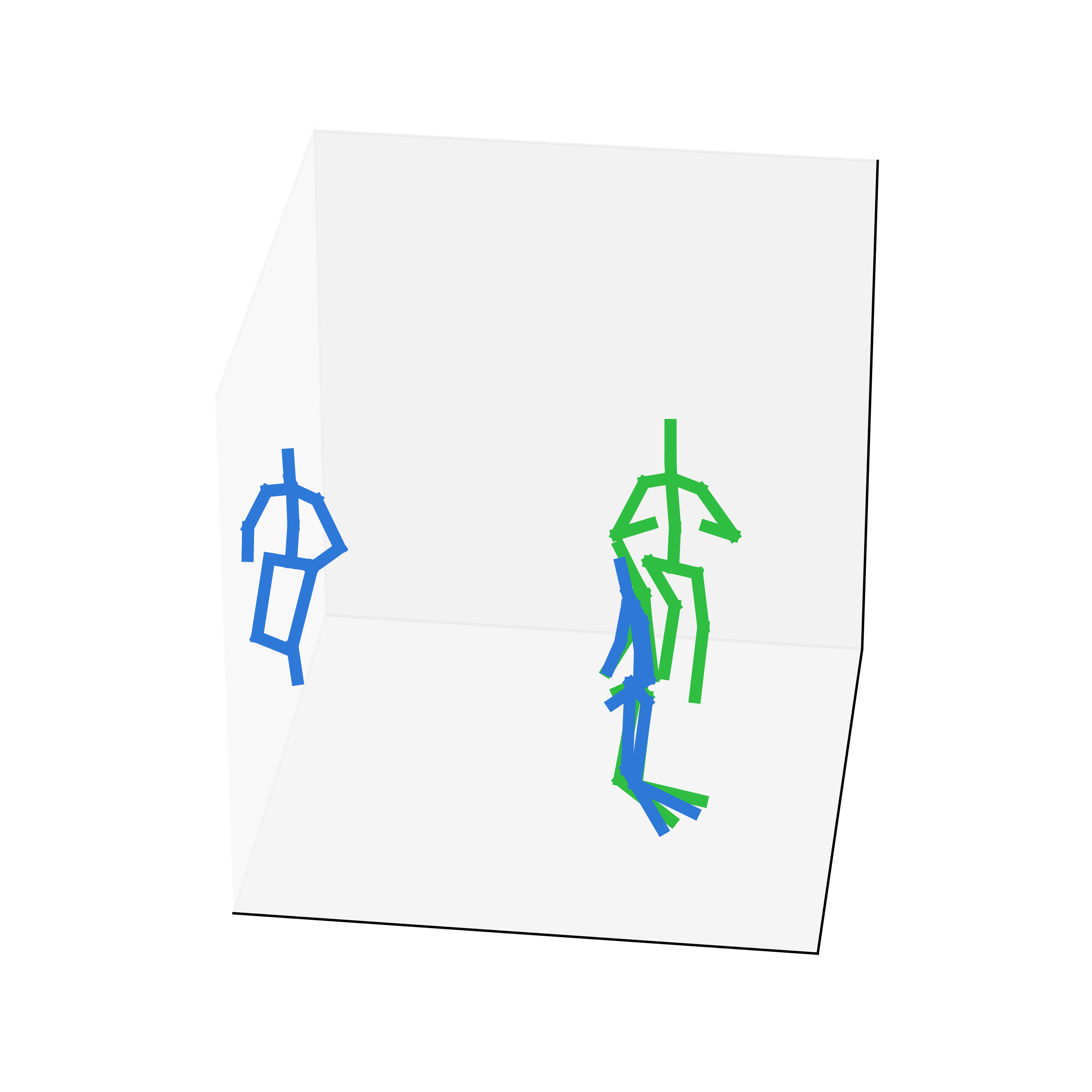} \end{minipage} \\
\end{tabular}
\end{center}
\caption{\textbf{Qualitative results on MuPoTS-3D.} Top row contains random images from the top 10 percentile, medium row contains those from the middle 10 percentile and bottom row from the worst 10 percentile. Green skeletons are the ground truth poses, blue ones are our predictions. Note that not every person in an image has ground truth annotation.}
\label{fig:qualitative-res}
\end{figure*}

We present example outputs of our model in Figure \ref{fig:qualitative-res}. Each row shows random examples from the best, median and worse deciles. A common failure case of our model are non-standing poses (in he middle of the bottom row).

\subsection{Ablation studies}
We investigate the effectiveness of our JointDepthNet in Table \ref{tbl:ablation}. We trained our network without JointDepthNet, only using full supervision. Our weak supervision improves on the absolute metrics. However, on relative metrics they remain unchanged. This finding is consistent on other databases, see Table \ref{tbl:panoptic}. We attribute this to two facts. First, depth images hold information only for visible joints, so the root-relative location of an occluded joint can be guessed only with a high uncertainty. On the other hand, even if large part of the body is occluded, the absolute distance from the camera can still be deduced. Second, the relative error of the Kinect is different in the two tasks. The depth sensor of the Kinect has an error of 1--2~cm \cite{wasenmuller2017kinecterror}. The $z$ coordinate of a relative pose varies mostly between -50~cm and 50~cm in the MuPoTS-3D dataset, while the absolute depth is between 200~cm and 700~cm. That is, the error of the Kinect in proportion to the target value is 4--14 times larger for relative pose estimation than for absolute pose estimation.

We also show the effect of Layer Normalization vs Batch Normalization. When using Batch Normalization instead of Layer Normalization, the performance drops considerably in all metrics.

\begin{table}[h]
\caption{\textbf{Ablation studies}. We turned off features of our network separately. All results are on MuPoTS-3D, using unnormalized coordinates. 3DPCK is calculated on detected poses. }
\label{tbl:ablation}
\begin{center}
\begin{tabular}{lcccc}

& A-MPJPE $\downarrow$ & A-3DPCK $\uparrow$ & R-MPJPE $\downarrow$ & R-3DPCK $\uparrow$   \\
\hline
BatchNorm vs LayerNorm & 288 & 28.1 & 114 & 75.2  \\  
w/o JDN & 264 & 33.0 & 108 & 78.7 \\ 
\hline

Full model & 255 & 35.9 & 108 & 78.7 \\ 
\end{tabular}
\end{center}
\end{table}

\section{Conclusion and future work}
We proposed a multi-person absolute pose estimation algorithm that can utilize unannotated RGB\nobreakdash-D datasets. The inclusion of depth images improved absolute pose metrics over two datasets. We also achieved new state-of-the-art results on the MuPoTS-3D dataset in absolute pose estimation, beating previous best results by a large margin. However, the weak supervision did not affect the relative pose estimation results. We attribute this to self-occlusion and measurement errors.

In future work, larger RGB\nobreakdash-D datasets can be explored. In our work we used only the Panoptic dataset but creating a large, unified database of RGB\nobreakdash-D images with human poses could bring further improvements. 

\section*{Acknowledgment}
 MV received support from the European Union and co-financed by the European Social Fund (EFOP-3.6.3-16-2017-00002). AL was supported by the National Research, Development and Innovation Fund of Hungary via the Thematic Excellence Programme funding scheme under Project no. ED\_18-1-2019-0030 titled Application-specific highly reliable IT solutions

\bibliographystyle{splncs04}
\bibliography{refs}

\begin{thebibliography}{10}
\providecommand{\url}[1]{\texttt{#1}}
\providecommand{\urlprefix}{URL }
\providecommand{\doi}[1]{https://doi.org/#1}

\bibitem{ba2016layer}
Ba, J.L., Kiros, J.R., Hinton, G.E.: Layer normalization. arXiv preprint,
  arXiv:1607.06450  (2016)

\bibitem{belagiannis2014pictorialpose}
{Belagiannis}, V., {Amin}, S., {Andriluka}, M., {Schiele}, B., {Navab}, N.,
  {Ilic}, S.: 3d pictorial structures for multiple human pose estimation. In:
  CVPR. pp. 1669--1676 (June 2014)

\bibitem{cai2018weakhand}
Cai, Y., Ge, L., Cai, J., Yuan, J.: Weakly-supervised 3d hand pose estimation
  from monocular rgb images. In: ECCV. pp. 678--694. Springer International
  Publishing, Cham (2018)

\bibitem{dabral2019multiperson}
Dabral, R., Gundavarapu, N.B., Mitra, R., Sharma, A., Ramakrishnan, G., Jain,
  A.: Multi-person 3d human pose estimation from monocular images. In: 3DV
  (2019)

\bibitem{drover_2dto3d}
Drover, D., M.~V, R., Chen, C.H., Agrawal, A., Tyagi, A., Huynh, C.P.: Can 3d
  pose be learned from 2d projections alone? ECCV Workshops p. 78–94 (2019)

\bibitem{fang2018posegrammar}
Fang, H.S., Xu, Y., Wang, W., Liu, X., Zhu, S.C.: Learning pose grammar to
  encode human body configuration for 3d pose estimation. AAAI  (2018)

\bibitem{firman2016rgbddatasets}
Firman, M.: Rgbd datasets: Past, present and future. CVPR Workshops  (Jun 2016)

\bibitem{geman-mcclure}
Geman, S., McClure, D.E.: Statistical methods for tomographic image
  reconstruction. Bulletin of the International Statistical Institut
  \textbf{52}(4),  5--21 (1987)

\bibitem{maskrcnn}
He, K., Gkioxari, G., Dollar, P., Girshick, R.: Mask {R-CNN}. ICCV  (Oct 2017)

\bibitem{Hossain2017temporal}
Hossain, M.R.I., Little, J.J.: Exploiting temporal information for 3d pose
  estimation (2017)

\bibitem{batchnorm}
Ioffe, S., Szegedy, C.: Batch normalization: Accelerating deep network training
  by reducing internal covariate shift. In: ICML. pp. 448--456 (2015)

\bibitem{h36m}
Ionescu, C., Papava, D., Olaru, V., Sminchisescu, C.: Human3.6m: Large scale
  datasets and predictive methods for 3d human sensing in natural environments.
  TPAMI  \textbf{36}(7),  1325--1339 (July 2014)

\bibitem{panoptic}
Joo, H., Liu, H., Tan, L., Gui, L., Nabbe, B., Matthews, I., Kanade, T.,
  Nobuhara, S., Sheikh, Y.: Panoptic studio: A massively multiview system for
  social motion capture. In: ICCV (2015)

\bibitem{kocabas2019epipolar}
Kocabas, M., Karagoz, S., Akbas, E.: Self-supervised learning of 3d human pose
  using multi-view geometry. In: CVPR (June 2019)

\bibitem{li2019mdn}
Li, C., Lee, G.H.: Generating multiple hypotheses for 3d human pose estimation
  with mixture density network. In: CVPR (June 2019)

\bibitem{lichan2015humanpose}
Li, S., Chan, A.B.: 3d human pose estimation from monocular images with deep
  convolutional neural network. In: Cremers, D., Reid, I., Saito, H., Yang,
  M.H. (eds.) ACCV. pp. 332--347. Springer International Publishing, Cham
  (2015)

\bibitem{megadepth2018}
Li, Z., Snavely, N.: Megadepth: Learning single-view depth prediction from
  internet photos. In: CVPR (2018)

\bibitem{3dbaseline}
Martinez, J., Hossain, R., Romero, J., Little, J.J.: A simple yet effective
  baseline for 3d human pose estimation. In: ICCV. pp. 2659--2668 (2017)

\bibitem{mehta}
Mehta, D., Rhodin, H., Casas, D., Fua, P., Sotnychenko, O., Xu, W., Theobalt,
  C.: Monocular 3d human pose estimation in the wild using improved {CNN}
  supervision. In: 3DV. pp. 506--516 (2017)

\bibitem{mehta2018single_shot}
Mehta, D., Sotnychenko, O., Mueller, F., Xu, W., Sridhar, S., Pons-Moll, G.,
  Theobalt, C.: Single-shot multi-person 3d pose estimation from monocular
  {RGB}. In: 3DV (sep 2018)

\bibitem{moon2019camdistance}
Moon, G., Chang, J.Y., Lee, K.M.: Towards 3d human pose estimation in the wild:
  A weakly-supervised approach. In: ICCV (Oct 2019)

\bibitem{pavlakos2018ordinal}
Pavlakos, G., Zhou, X., Daniilidis, K.: Ordinal depth supervision for 3d human
  pose estimation. In: CVPR (2018)

\bibitem{gorog}
Pavlakos, G., Zhou, X., Derpanis, K.G., Daniilidis, K.: Coarse-to-fine
  volumetric prediction for single-image 3d human pose. In: CVPR. pp.
  1263--1272 (2017)

\bibitem{ramakrishna2012dictionary}
Ramakrishna, V., Kanade, T., Sheikh, Y.: Reconstructing 3d human pose from 2d
  image landmarks. In: ECCV. pp. 573--586 (2012)

\bibitem{helge_geometry-aware}
Rhodin, H., Salzmann, M., Fua, P.: Unsupervised geometry-aware representation
  for 3d human pose estimation. In: ECCV (2018)

\bibitem{rogez2017lcrnet}
Rogez, G., Weinzaepfel, P., Schmid, C.: {LCR-N}et:
  Localization-classification-regression for human pose. In: CVPR. pp.
  1216--1224 (July 2017)

\bibitem{rogez2019lcrpp}
Rogez, G., Weinzaepfel, P., Schmid, C.: Lcr-net++: Multi-person 2d and 3d pose
  detection in natural images. TPAMI  (2019)

\bibitem{humaneva}
Sigal, L., Balan, A.O., Black, M.J.: Human{E}va: Synchronized video and motion
  capture dataset and baseline algorithm for evaluation of articulated human
  motion. IJCV  \textbf{87}(1), ~4 (Aug 2009)

\bibitem{hrnet}
Sun, K., Xiao, B., Liu, D., Wang, J.: Deep high-resolution representation
  learning for human pose estimation. In: CVPR (2019)

\bibitem{integralPose}
Sun, X., Xiao, B., Liang, S., Wei, Y.: Integral human pose regression. In:
  ECCV. pp. 529--545 (September 2018)

\bibitem{veges2019depthpose}
{V\'eges}, M., {L{\H o}rincz}, A.: Absolute human pose estimation with depth
  prediction network. In: IJCNN. pp.~1--7 (July 2019)

\bibitem{veges2018siamese}
V\'eges, M., Varga, V., L{\H o}rincz, A.: 3d human pose estimation with siamese
  equivariant embedding. Neurocomputing  \textbf{339},  194 -- 201 (2019)

\bibitem{wan2019selfsupervised-hand}
Wan, C., Probst, T., Gool, L.V., Yao, A.: Self-supervised 3d hand pose
  estimation through training by fitting. In: CVPR. pp. 10853--10862 (2019)

\bibitem{llcs}
Wang, L., Chen, Y., Guo, Z., Qian, K., Lin, M., Li, H., Ren, J.S.: Generalizing
  monocular 3d human pose estimation in the wild. In: ICCV Workshops (Oct 2019)

\bibitem{wasenmuller2017kinecterror}
Wasenm{\"u}ller, O., Stricker, D.: Comparison of {K}inect {V}1 and {V}2 depth
  images in terms of accuracy and precision. In: ACCV Workshops. pp. 34--45.
  Cham (2017)

\bibitem{yuan2018privilegedhand}
Yuan, S., Stenger, B., Kim, T.K.: Rgb-based 3d hand pose estimation via
  privileged learning with depth images (2018)

\bibitem{zanfir2018smpl3dpose}
Zanfir, A., Marinoiu, E., Sminchisescu, C.: Monocular 3d pose and shape
  estimation of multiple people in natural scenes the importance of multiple
  scene constraints. CVPR pp. 2148--2157 (2018)

\bibitem{zanfir2018multipose}
Zanfir, A., Marinoiu, E., Zanfir, M., Popa, A.I., Sminchisescu, C.: Deep
  network for the integrated 3d sensing of multiple people in natural images.
  In: NIPS, pp. 8410--8419 (2018)

\bibitem{zhou2015dict}
Zhou, X., Leonardos, S., Hu, X., Daniilidis, K.: 3d shape estimation from 2d
  landmarks: A convex relaxation approach. In: CVPR. pp. 4447--4455 (2015)

\bibitem{zhou2017}
Zhou, X., Huang, Q., Sun, X., Xue, X., Wei, Y.: Towards 3d human pose
  estimation in the wild: A weakly-supervised approach. In: ICCV. pp. 398--407
  (Oct 2017)

\end{thebibliography}

\end{document}